\documentclass[pre, twocolumn, amsmath, amssymb, superscriptaddress]{revtex4}

\usepackage{comment}
\usepackage{graphicx}
\usepackage{dcolumn}
\usepackage{bm}
\usepackage{braket}
\usepackage{xcolor}
\usepackage{hyperref}
\usepackage{amsmath}
\usepackage{siunitx}
\usepackage{xspace}

\newcommand{\beq}{\begin{equation}}
\newcommand{\eeq}{\end{equation}}

\usepackage{mathtools}

\newcommand{\eqname}{{Eq.}\xspace}

\newcounter{supportingfigure}

\begin{document}

\title{A new pathway to generative artificial intelligence\\ by minimizing the maximum entropy}

\author{Mattia Miotto}
\affiliation{Center for Life Nanoscience, Istituto Italiano di Tecnologia, Viale Regina Elena 291,  00161, Rome, Italy}

\author{Lorenzo Monacelli}
\affiliation{Department of Physics, Sapienza University, Piazzale Aldo Moro 5, 00185, Rome, Italy}

\begin{abstract}
Generative artificial intelligence revolutionized society. Current models are trained by minimizing the distance between the produced data and the training set. Consequently, development is plateauing as they are intrinsically data-hungry and challenging to direct during the generative process.  
To overcome these limitations, we introduce a paradigm shift through a framework where we do not fit the training set but find the most informative yet least noisy representation of the data -- simultaneously minimizing the entropy to reduce noise and maximizing it to remain unbiased via adversary training. 

The result is a general physics-driven model, which is data-efficient and flexible, permitting to control and influence the generative process. Benchmarking shows that our approach outperforms variational autoencoders. We demonstrate the method’s effectiveness in generating images, even with limited training data, and its unprecedented capability to customize the generation process a posteriori without any fine-tuning or retraining.
\end{abstract}

\maketitle



Generative artificial intelligence (GenAI) refers to models capable of algorithmically producing -- novel -- data resembling those from the training set. 
Text generative models, for instance, predict the probability of each possible next token, \textit{i.e.} clusters of characters, of a sequence to generate a plausible continuation of an initial prompt~\cite{Lv2023}. 

Multiple algorithms have been developed for such tasks, each offering distinct advantages depending on the data type. For instance, transformers are particularly effective for sequence generation, as seen in large language models \cite{vaswani_attention_2023}, while Generative Adversarial Networks (GANs) \cite{Goodfellow2020}, Variational Autoencoders (VAEs) \cite{kingma_auto-encoding_2022}, and Diffusion models \cite{diffmodel} are well-suited for handling multidimensional data, such as images.
Thanks to these models/architectures, GenAI is being used to address a wide range of complex problems~\cite{Gupta2024}, from designing drugs~\cite{Wu2024, Gangwal2024} and functional proteins\cite{Trinquier2021} to the discovery of novel materials~\cite{Zeni2025}, from advertising and entertainment~\cite{Totlani2023} to  education~\cite{Law2024} and communication~\cite{Roumeliotis2023}. 

\begin{figure}
    \centering
    \includegraphics[width=0.9\linewidth]{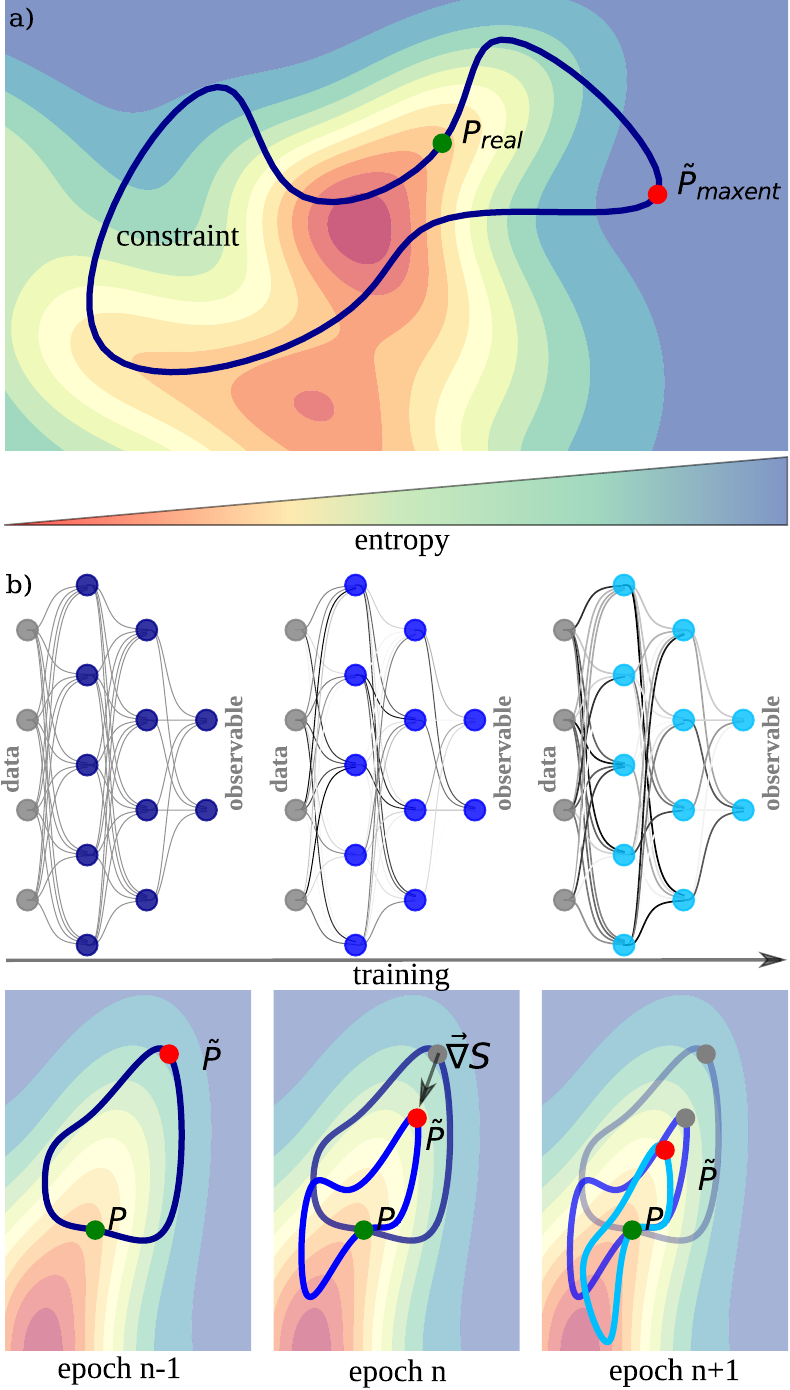}
    \caption{ \textbf{Illustration of the minimal maximum entropy principle.} \textbf{a)} Two dimensional representation of a possible  entropy landscape as a function of two reaction coordinates, with a dark blue line marking the values of the entropy subject to a given constraint/observable. Colors shift from red to blue as the entropy increases.   Green and red dots identify the real and maximum entropies values, respectively.  
    \textbf{b)} The observables/constrains can be defined in an unsupervised way using a neural network to extract the relevant features from the input data. Starting from a fully connected network with equal weights, the minimal maximum entropy algorithm adjusts the network parameters to obtain the constrain whose associated maximum entropy is minimum, i.e. closer to the real entropy. }
    \label{fig:1}
\end{figure}

Parallel to the enormous spreading of GenAI, serious ethical concerns regarding the generated content are being formulated~\cite{carlini2023}, and the feasibility of further improving by just scaling existing architectures is questioned~\cite{Jones2024}. Indeed, the extremely data-greedy nature of most GenAI models is leading to a saturation of the available data~\cite{Jones2024}, while training with generated samples is demonstrated to poison the models~\cite{Shumailov2024, poison_data}.  
While these models can rapidly generate novel samples, customizing the results remains challenging, often requiring multiple, supervised random attempts to steer the outcome \cite{li_controlnet_2024,huang2023composer}. Considerable efforts have been revolved to mitigate this issue~\cite{zhang_adding_2023,li_controlnet_2024,hu_lora_2021,xie2023smartbrush,huang2023composer,rombach2022high}, that still relies on model-specific solutions, requiring \textit{ad hoc} retraining each time the underlying model is updated. 


In this work, we propose a different route for GenAI based on the 
`minimal maximum entropy' principle. We demonstrate that this approach is robust in presence of under-sampled data and readily customizable, allowing the generative process to be directed \emph{a posteriori} without requiring retraining.

\begin{figure*}
    \centering
    \includegraphics[width=0.99\linewidth]{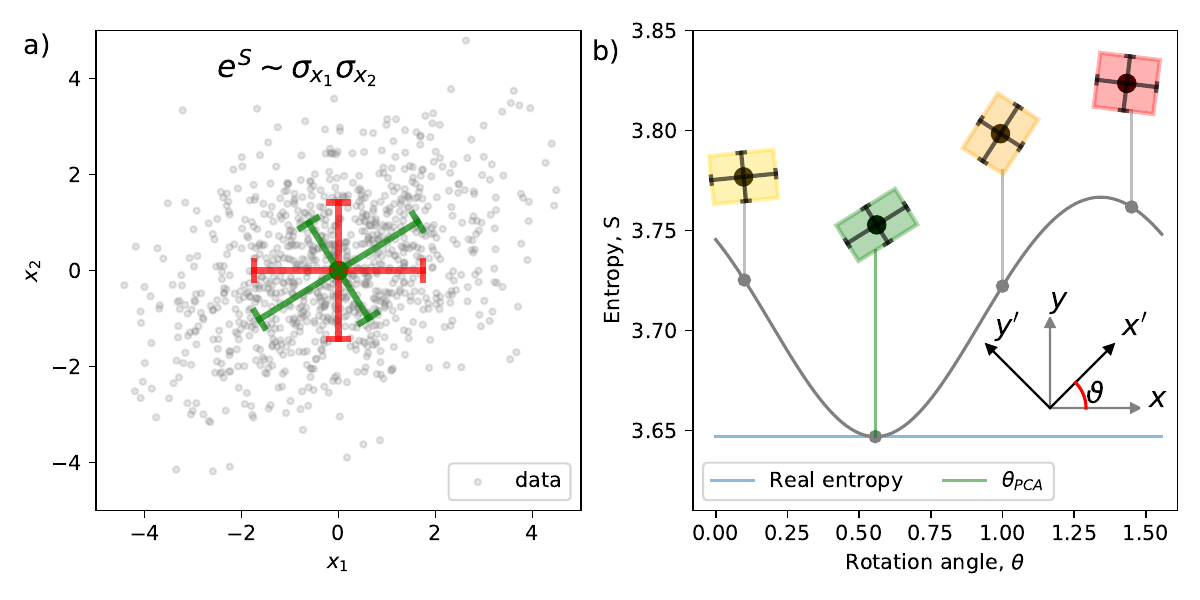}
    \caption{\textbf{Principal component analysis as a special Min-MaxEnt solution.} \textbf{a)} Example of 2D normal distributed data. Red and green bars represent the squared roots of the variances in two different basis. \textbf{b)} Entropy as a function of the rotation angle of the basis in which data are represented.}
    \label{fig:2}
\end{figure*}

\section{The minimal maximum entropy principle}

Maximum entropy is a guiding principle to assign probabilities to events~\cite{Bialek2012-lp}. Indeed,  maximizing entropy selects the most unbiased probability distribution consistent with given constraints, ensuring no unwarranted assumptions are made beyond the available information. Given a set of measures $f_i(x)$, the probability distribution that (i) maximizes entropy and (ii) ensures that the expected values of $f_i$ match those observed in the training set, is given by:
\begin{equation}
    \tilde P_{f_i,\lambda_i}(x) = \frac{1}{Z}\exp\left[-\sum_i \lambda_i f_i(x)\right],
    \label{eq:maxent}
\end{equation}
with $Z = \int dx \exp\left[-\sum_i \lambda_i f_i(x)\right]$. Here, the integral marginalize over all possible configurations $x$ and $\lambda_i$ are Lagrange multipliers that constrain the average values of the measures $f_i(x)$ to match those of the training set.
\begin{equation}
    \int dx f_i(x) \frac{1}{Z}\exp\left[-\sum_i \lambda_i f_i(x)\right] = \frac{1}{N}\sum_{\{x\}_{\text{train set}}} f_i(x) = \mu_i.
    \label{eq:constrain}
\end{equation}

Operationally, training a maximum entropy model consists of the following steps: i) define a set of measures $f_i(x)$, ii) compute the average values $\mu_i$ of such measures on the training dataset, iii) solve iteratively \eqname~\eqref{eq:constrain} to find the values of $\lambda_i$.
Once we got the converged $\lambda_i$ values, the maximum entropy principle ensures that the entropy of the target $\tilde P_{f_i, \lambda_i}(x)$ is always above the exact entropy~\cite{ecolat}. Since both the real probability distribution, $P(x)$, and the MaxEnt one $\tilde P_{f_i, \lambda_i}(x)$ satisfy the constraints 
$ \int dx f_i(x) P(x) = \mu_i$, they belong to the same manifold of all the distributions satisfying $\left<f_i(x)\right> = \mu_i$, where $\braket{\cdot}$ indicates the expected value (sketched in Figure~\ref{fig:1}\textbf{a}). $\tilde P_{f_i,\lambda_i}(x)$ maximizes the entropy within this manifold, thus, $S[\tilde P_{f_i,\lambda_i}] $ is always higher or equal to the real distribution Shannon's entropy $S[P]$: $S[\tilde P_{f_i,\lambda_i}] \ge S[P]$. 

The inequality between real and MaxEnt entropy sets up a  variational principle: for a fixed set of measures $f_i$, the entropy of the corresponding MaxEnt distribution is always above the real one. The residual entropy difference quantifies how much information can be further extracted from the data by improving the choice of $f_i(x)$. 
From the minimal maximum entropy principle, we can introduce a Min-MaxEnt algorithm to optimize the set of measures $f_i$ by minimizing the entropy of the MaxEnt distribution $S[\tilde P{f_i, \lambda_i}]$:
\begin{equation}
    S[P] = \min_{f_i} S[\tilde P_{f_i,\lambda}]
    \label{eq:minmaxent:opt}
\end{equation}
The equality in \eqname~\eqref{eq:minmaxent:opt} holds as it always exists a set of measures $f_i$ that uniquely define a probability distribution~\cite{ecolat}.

This idea was first introduced almost thirty years ago~\cite{6796444}, but its employment has been dampened by the challenge of evaluating the entropy of the obtained MaxEnt distribution. This limited the application only to observables $f_i(x)$ for which an  analytical or mean-field expressions of the partition function is computable~\cite{Lynn2025}.  
Here, we solve this issue deriving an exact expression for the functional gradient of the MaxEnt entropy, allowing for the application of the Min-MaxEnt for any choice of   $f_i(x)$, including deep neural networks.

In practice, the measures $f_i$ can be parametrized as generic nonlinear functions of the configurations, depending on a vector of parameters $\theta_1, \cdots, \theta_n$ (see Figure~\ref{fig:1}), and the minimization of the entropy can be performed directly optimizing $\theta_1, \cdots, \theta_n$:
\begin{equation}
    \tilde P\left[\{\theta_i\}_1^n, \{\lambda_i\}_1^n\right](x) = \frac{1}{Z} \exp\left[-\sum_i \lambda_i f_i[\theta_1, \cdots, \theta_n](x)\right]
\end{equation}

\begin{figure*}
    \centering
    \includegraphics[width=1.\linewidth]{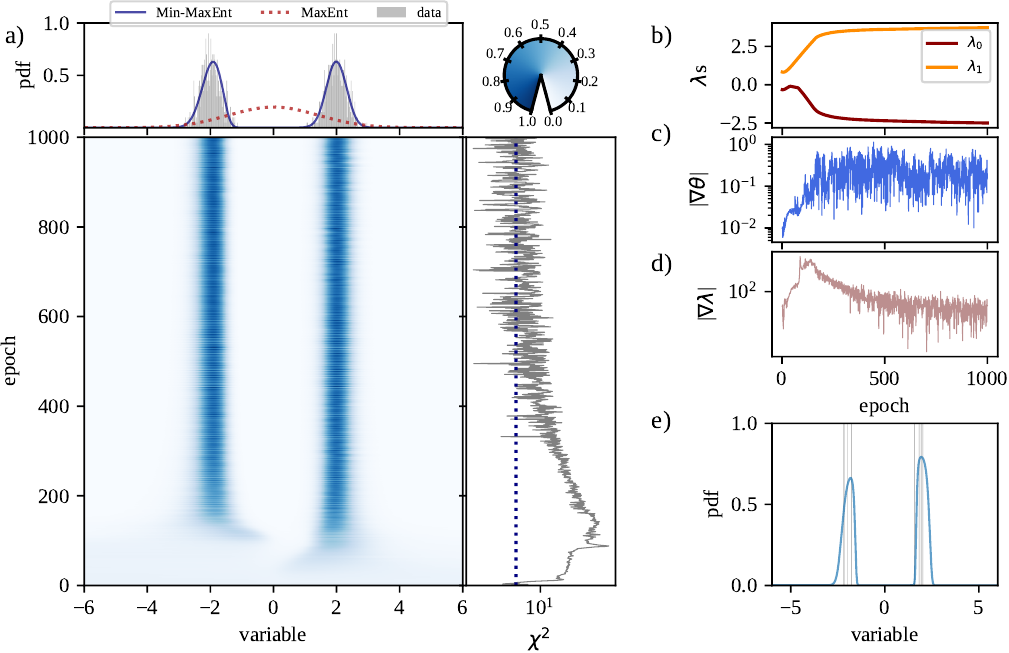}
    \caption{\textbf{Inference of a bimodal normal distribution.} \textbf{a)} Inferred Min-MaxEnt distribution as a function of the training epochs for a dataset of 1000  bimodal normal variables. In the top panel, the best-inferred MaxEnt and Min-MaxEnt distributions are reported in red and blue, respectively. Real data distribution is shown in gray. \textbf{b)} Values of the Lagrangian multipliers as a function of the training epochs. \textbf{c)} Modulus of the gradient of the observables and \textbf{d)} Lagrangian multipliers as a function of the training epochs. \textbf{e)} Best inferred  Min-MaxEnt distributions and real data distribution for a training performed with a dataset composed of 10 samples. }
    \label{fig:3}
\end{figure*}

Therefore, we must simultaneously train the $\lambda_i$ parameters to constrain the averages of $f_i$ (\eqname~\ref{eq:constrain}), and the $\theta_i$ parameters to minimize the maximum entropy (\eqname~\ref{eq:minmaxent:opt}).
To optimize $\lambda_i$, we define a cost function quantifying the displacement of the $f_i$ averages between training and generated samples~\cite{ecolat, miotto_tolomeo_2021}: 
\begin{equation}
    \chi^2 = \sum_i\frac{(\braket{f_i} - \mu_i)^2}{\sigma_i^2} 
    \label{eq:chi2}
\end{equation}
where $\sigma_i^2$ is the variance of the $i$-th observable on the training dataset. The minimization of \eqname~\eqref{eq:chi2} can be pre-conditioned as discussed in Ref. \cite{ecolat}. The parameters $\lambda_i$ are updated with a gradient descend algorithm according to
\begin{equation}
    \lambda_i \longrightarrow \lambda_i - \eta \frac{\partial \chi^2}{\partial\lambda_i}
    \label{eq:update:lambda}
\end{equation}
(In this work, we employed the ADAM algorithm~\cite{ADAM}).
Next, we introduce an update rule also for the $\theta_i$ parameters that decrease the maximum entropy. 
As sketched in Figure~\ref{fig:1}b, updating $\theta_1\cdots\theta_n$  progressively modifies the constraint manifold, minimizing the entropy of the corresponding MaxEnt distribution.
This is achieved by computing the gradient of the MaxEnt distribution's entropy
\begin{equation}
    \theta_i \longrightarrow \theta_i - \eta \frac{dS}{d\theta_i} 
    \label{eq:update:theta}.
\end{equation}

Notably, the entropy itself is practically incomputable. However, the entropy gradient is a standard observable and can be evaluated efficiently as:
\begin{equation}
    \frac{dS}{d\theta_i} = - \sum_j \lambda_j \left(\left<\frac{d f_j}{d\theta_j}\right>_{\tilde P[\theta_1, \cdots,\theta_n,\lambda_1, \cdots,\lambda_n]} - \left<\frac{d f_j}{d\theta_j}\right>_{P}\right),
    \label{eq:minim:entropy}
\end{equation}
where $\left<\frac{d f_j}{d\theta_j}\right>_{\tilde P[\theta_1, \cdots,\theta_n,\lambda_1, \cdots,\lambda_n]}$ is obtained by averaging an ensemble of configurations generated with the current MaxEnt distribution, while $\left<\frac{d f_j}{d\theta_j}\right>_{P}$ is evaluated on the real distribution, i.e., the training set. The formal proof of \eqname~\eqref{eq:minim:entropy} is reported in Supplementary Materials. 
\eqname~\eqref{eq:minim:entropy} can be implemented with the usual backpropagation by defining an auxiliary cost function $\tilde S$
\begin{multline}
    \tilde S(\theta_1,\cdots,\theta_n) = \frac{1}{N_1}\sum_{i = 1}^{N_1} \lambda_i f_i[\theta_1,\cdots,\theta_n](x_i) - \\ -  \frac{1}{N_2}\sum_{i = 1}^{N_2} \lambda_i f_i[\theta_1,\cdots,\theta_n](\tilde x_i), 
\end{multline}
\begin{equation}
    \frac{dS}{d\theta_i} = \frac{\partial\tilde S}{\partial\theta_i},
\end{equation}
where the configurations $\tilde x_i$ are sampled through the MaxEnt probability distribution. 
The gradient of $\tilde S$ does not depend explicitly on the probability distribution, therefore, the backpropagation is fast as it does not require running through the ensemble generation.  
Note that the double optimization of all the $\theta_i$ and $\lambda_i$ parameters works like an adversary competition: the $\lambda_i$ optimization aims at maximizing entropy with the given set of constraints, while the $\theta_i$ optimization alters the set of constraints to minimize the entropy of the distribution, extracting order from disorder. 
Unlike most machine-learning approaches, the optimization rule in \eqname~\eqref{eq:minim:entropy} does not evaluate a distance between the generated data and the training, thus mitigating the risks of overfitting.

In the following sections, we discuss different applications of our approach. First, we focus on a special case where analytical insight can be gained.  In particular, we show that Principal component analysis (PCA) can be formally derived from the Min-MaxEnt principle. Next, we probe the capability of the Min-MaxEnt to infer different kinds of 1D bimodal distributions against the predictions of standard MaxEnt 
and variational autoencoders (VAE). Finally, we apply the method to the contest of image generation,
 demonstrating (i) its
capabilities when trained on a small subset of data,  (ii) how it can be refined via adversary network training, and (iii) how controlled generation can be easily enforced on the trained model.

\begin{figure*}
    \centering
    \includegraphics[width=0.99\linewidth]{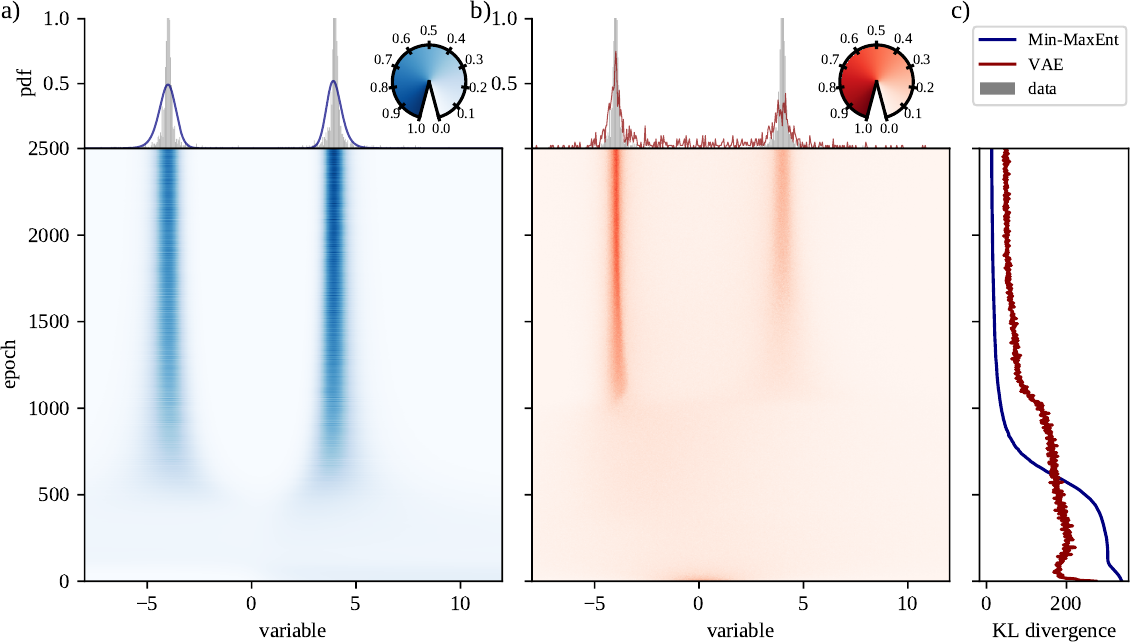}
    \caption{\textbf{Min-MaxEnt vs variational auto-encoder.} \textbf{a)} Inferred Min-MaxEnt distribution as a function of the training epochs for a dataset of 1000  bimodal Lorentians variables. In the top panel, the best-inferred Min-MaxEnt distribution is reported blue, while real data distribution is shown in gray. \textbf{b)} Inferred VAE distribution as a function of the training epochs for a dataset of 2000  bimodal Lorentians variables. In the top panel, 
    the best-inferred VAE distribution is reported red, while real data distribution is shown in gray. \textbf{c)} Kullback–Leibler divergence between real and inferred distributions as a function of the training epochs for the Min-MaxEnt and VAE methods.}
    \label{fig:4}
\end{figure*}

\begin{figure*}
    \centering
    \includegraphics[width=0.9\linewidth]{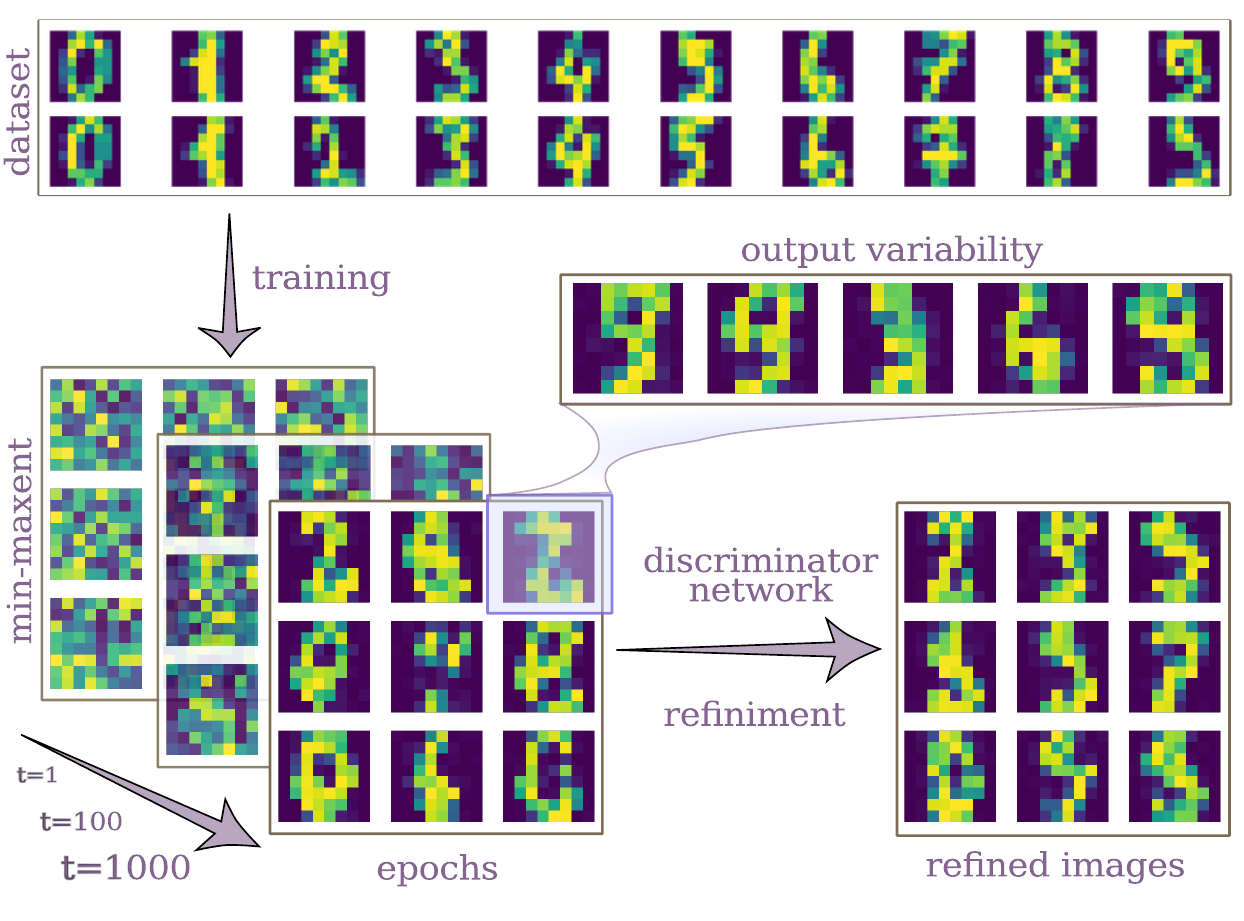}
    \caption{ \textbf{Image generation.} Example of the generative power of the proposed Min-MaxEnt principle in the case of images. Starting from the 8x8 MNIST dataset, a Min-MaxEnt is trained using a deep neural network with 16 output observable. The results can be further refined using a discriminator network to bias the generation process. }
    \label{fig:5}
\end{figure*}

\begin{figure}
    \centering
    \includegraphics[width=0.99\linewidth]{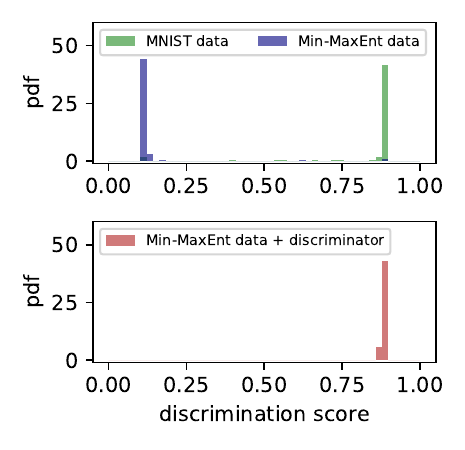}
    \caption{ \textbf{Discriminator procedure.} \textbf{a)} Probability distribution of the post-training discrimination score for MNIST data (real) and Min-MaxEnt data (generated). 
    \textbf{b)} Probability distribution of the post-training discrimination score for data generated via the Min-MaxEnt algorithm with the addition of the discriminator bias.}
    \label{fig:gan}
\end{figure}

\begin{figure*}
    \centering
    \includegraphics[width=\linewidth]{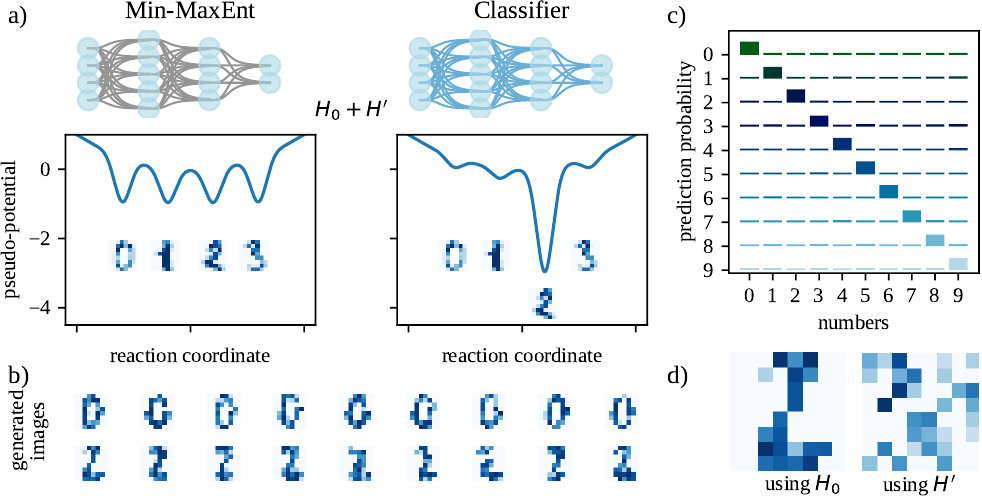}
    \caption{\textbf{Example of preferential-generation process.} \textbf{a)} Pictorial representation of the Min-MaxEnt process in MNIST dataset: pseudo-hamiltonian, $H_0$, can be interpreted as a pseudo-potential whose minima correspond to different kinds of configurations of the system, i.e., numbers.  Adding a classifier neural network as a perturbation $H'$ to $H_0$  alters the probability of sampling certain kinds of configurations. \textbf{b)} Example of biased generations of 0 (top row) and 2 (bottom row). \textbf{c)} Classifier capacity of assigning a given configuration to numbers from 0 to 9.
    \textbf{d)} Comparison between a configuration obtained via a Monte Carlo process using only $H_0$ vs using only $H'$.}
    \label{fig:6}
\end{figure*}

\subsection{Principal component analysis as a special Min-MaxEnt solution}

Principal component analysis is a widely used statistical approach to represent high-dimensional data in its essential features or principal components. Such components are obtained by the linear combinations of the original variables that diagonalize the covariance matrix. 
Due to its capability to reduce dimensionality by retaining only the components with the highest variances, PCA finds wide applications, especially in fields characterized by the presence of vast amounts of data, from bioinformatics~\cite{Ma2011} to particle physics~\cite{Altsybeev2020}, in tasks like estimating missing values in huge data matrices, sparse component estimation, and the analysis of images, shapes, and functions~\cite{Greenacre2022}. 

In the following, we demonstrate that PCA can be retrieved as a particular solution of the minimal maximum entropy principle by constraining the variance of an arbitrary linear combination of the system variables.     
Figure~\ref{fig:2}a) shows a straightforward 2D case in which data (gray dots) are drawn from a general probability distribution with a covariance matrix $\Sigma_{ij} = \left<x_ix_j\right>_P$ of elements $\Sigma_{11}=3$, $\Sigma_{22}=2$, and $\Sigma_{12}=1$. 

To constrain the variances of a linear combination of the variables, we define the $f_i[\theta](x, y)$ observables as
\begin{equation}
\left\{\begin{array}{l}
f_1(x) = (\cos\theta x + \sin\theta y)^2 \\ 
f_2(x) = (-\sin\theta x + \cos\theta y)^2
\end{array}\right. .
\label{eq:pca2:obs}
\end{equation}
The corresponding MaxEnt solution is a normal distribution of the form
\begin{align}
    \tilde P[\theta](x,y) \propto \exp\bigg[&\lambda_1 \left(\cos\theta x + \sin\theta y\right)^2 + \nonumber \\ 
    &\lambda_2\left(-\sin\theta x + \cos\theta y\right)^2\bigg],
\end{align}
where the $\lambda_i$ can be found analytically imposing that the averages of $f_i(x)$ matches with the exact distribution:
\begin{eqnarray}
    \tilde\Sigma_{11} = \frac{1}{2\lambda_1} = \Sigma_{11}\cos^2\theta + \sin^2\theta \Sigma_{22} + \Sigma_{12}\sin2\theta,\\
    \tilde\Sigma_{22} = \frac{1}{2\lambda_2} = \Sigma_{11}\sin^2\theta + \cos^2\theta \Sigma_{22} - \Sigma_{12}\sin2\theta.
\end{eqnarray}
The entropy of this MaxEnt distribution (Gaussian) is analytical and only depends on the determinant of its covariance matrix, i.e., the product $\tilde\Sigma_{11}\tilde\Sigma_{22}$
\begin{equation}
    S  = \frac{1}{2} \ln\left(4\pi^2 e^2\tilde\Sigma_{11}\tilde\Sigma_{22} \right) .
\end{equation}
Figure~\ref{fig:2}b displays the entropy as a function of the rotation angle $\theta$. 
Minimizing $S$ is equivalent to minimizing the product $\tilde\Sigma_{11}\tilde\Sigma_{22}$. It can be shown that

\begin{multline}
\tilde\Sigma_{11}\tilde\Sigma_{22} = \left(\frac{\sin 2\theta}{2}[\Sigma_{11} -\Sigma_{22}] -  \cos 2\theta  \Sigma_{12}\right)^2 +  \text{const},
\label{eq:}
\end{multline}
which is trivially minimized for
\begin{equation}
\theta^\star = \frac{1}{2} \arctan\left(\frac{2~\Sigma_{12}}{\Sigma_{11}-\Sigma_{22}}\right)   \label{eq:angle}
\end{equation}

The angle in \eqname~\eqref{eq:angle} also identifies the rotation diagonalizing the real covariance matrix $\Sigma$, i.e. the PCA solution. In fact, diagonalizing $\sigma$, requires that:
\begin{equation}
\left<(\cos\theta^\star x + \sin\theta^\star y)(\cos\theta^\star y - \sin\theta^\star x)\right>_P = 0.
\end{equation}
This establishes that the Min-MaxEnt distribution coincides with the PCA solution when the observables $f_i$ are defined as in \eqname~\eqref{eq:pca2:obs}. In the Supplementary Materials, we generalize this result to arbitrary dimensions, proving that the optimal PCA rotation emerges naturally from minimizing the MaxEnt entropy while constraining variances along an arbitrary basis.

\subsection{Neural network as Min-MaxEnt observables}

It is generally challenging to capture complex data patterns with an \emph{ad hoc} parametrization of the observables $f_i(x)$. In this section, we implement the $f_i(x)$ measures as the output layer of a neural network, which parametrizes a general nonlinear function. Specifically, we test the method's performance on a dataset generated from one-dimensional bimodal distributions.
First, we draw a set of 1000 training data (see the gray histogram in the top panel Figure~\ref{fig:3}a) from a bimodal distribution $P = \frac{1}{2}\mathcal{N}(x, \bar{x}_1, \sigma_1) + \frac{1}{2}\mathcal{N}(x, \bar{x}_2, \sigma_2)$, where $\mathcal{N}$ is a normal distribution with mean $\bar x$ and variance $\sigma^2$. 
Figure~\ref{fig:3} shows the results for a Min-MaxEnt run using two observables obtained as output nodes of a multilayer perceptron. Figure~\ref{fig:3}a displays the evolution of the predicted Min-MaxEnt probability distribution as the training epochs increase. The top panel compares the real data, the Min-MaxEnt distribution after 1000 epochs, and the MaxEnt distribution constraining mean and variance of $x$. After a few hundred epochs, the Min-MaxEnt distribution perfectly captures both the bimodality and the variances of the single peaks. 
Figure~\ref{fig:3}e displays the inferred Min-MaxEnt distribution (blue) trained only on 10 data points (gray). 

A more challenging test is to replace the normal distributions with a Cauchy (Lorentzian) distribution, which allows for under-sampled data, i.e. rare events, to occur far from the distribution peaks. In this case, we compare the Min-MaxEnt with a Variational Autoencoder where the Encoder network has the same architecture as the $f_i$ parametrization (see Figure~\ref{fig:4}).
The 1000 training data are drawn from the distribution $P(x) = \frac{1}{2}\mathcal{L}(x, {x}_0^1, \lambda_1) + \frac{1}{2}\mathcal{L}(x, {x}^2_0, \lambda_2)$, where $\mathcal{L}$ is a Cauchy distribution with median $x_0$ and half width at half maximum (HWHM) $\lambda$. Figure~\ref{fig:4}a and Figure~\ref{fig:4}b) display the training of the Min-MaxEnt and VAE, respectively. 
The final distributions are shown in the top panels with the real data. 
To quantitatively measure the difference between predicted and real distributions, in Figure~\ref{fig:4}c, we reported the Kullback-Leibler divergence between real and inferred distributions as a function of the training epochs. The better result of the Min-MaxEnt reflects the tendency of the VAE to overfit rare events, producing a noisy background and overestimating the probability in the regions far from the distribution peaks. This accounts for the notorious issue of VAE, which suffers from blurry generated samples compared to the data they have been trained on~\cite{Bredell2023}.

\subsection{Min-MaxEnt for image generation}

Next, we applied the Min-MaxEnt algorithm to the case of image generation,  using the MNIST dataset~\cite{mnist}, a collection of 1797 images of greyscale labeled handwritten digits, that are represented as 8x8 pixel matrices. We selected a training set of 200 images, discarding the labels to train the model. The observables are defined through a convolutional neural network (CNN) made by two convolutional layers, tailed by a last fully connected ending in 16 output nodes. The images are generated according to the MaxEnt probability distribution using a Metropolis-Monte Carlo Markov's chain~\cite{Hastings1970,ecolat,miotto_tolomeo_2021}.

The entropy reduction is evident from \figurename~\ref{fig:5}, where the generative process evolves from noisy images at the first epochs (high entropy) to more defined outputs. 
Unlike all other generative algorithms, the training procedures never enforce the model's output to replicate the training set. Therefore, no memory of specific images are stored within the network, preventing overfit and helping with generalization.

Once the model is trained, we have a final probability distribution from which images can be extracted. Consequently, it is possible to define an effective energy landscape for images, which can be further used to direct the generation process:
\begin{equation}
H_0(x) = \sum_{i=1}^{16} f_i[\theta_1\cdots\theta_n](x)\lambda_i.
\end{equation}
Similarly to physical systems, generation can be directed by adding an \emph{external field} $H'(x)$. 
For instance, we can train an independent network $g(x)$ to recognize generated data like  
\begin{equation}
    g(x) = \left\{\begin{array}{lr}
    1 & \text{generated image} \\
    0 & \text{real image}
    \end{array}\right.
\end{equation}
By altering the energy landscape as
\begin{equation}
    H(x) = H_0(x) + \underbrace{\alpha g(x)}_{H'(x)},
    \label{eq:gan}
\end{equation}
we can extract biased samples that are guaranteed to be indistinguishable from the original training set according to the network $g(x)$ (see Figure~\ref{fig:gan}). This process can be repeated by updating the training set of the discriminator to include some of the data generated with $H(x)$. 
Such an approach resembles the adversary network training, which can be applied efficiently to the Min-MaxEnt. In fact, the only network that needs to be re-trained is the discriminator since \eqname~\eqref{eq:gan} automatically generates indistinguishable samples for the respective classifier.

The effective energy landscape framework is also helpful for biasing the generation process toward specific targets. In most GenAI models, this involves some retraining of the network. In contrast, the generation through the Min-MaxEnt algorithm can be conditioned by introducing an external field modeled via a simple CNN classifier. For example, we trained a CNN classifier $h_i(x)$ to guess the labels encoding the written digit of the MNIST dataset (a task extremely easy for networks) like 
\begin{equation}
    h_i(x) = \left\{\begin{array}{lr}
         1 & \text{if } x\text{ represents the number }i \\
         0&\text{otherwise} 
    \end{array}\right.
\end{equation}
The biased generation is performed with the new energy landscape as
\begin{equation}
    H(x) = H_0(x) \underbrace{-\alpha  h_j(x) + \alpha \sum_{i\neq j}  h_j(x)}_{H'(x)}.
\end{equation}
The external field $H'(x)$ favors the generation of images that $h(x)$ classifies as the $j$ number. The result is shown in \figurename~\ref{fig:6}.
The classifier increases the potential energy around numbers different from the target and decreases it around the target. Interestingly, numbers generated via this method appear more readable, as images that cannot be clearly classified as one of the digits are un-favored. The training of $h(x)$ is completely independent of the Min-MaxEnt training, as the only training set employed is the original dataset of real images.
Panel d of \figurename~\ref{fig:6} shows what happens if we turn off the Min-Max-Ent Hamiltonian, and generate only according to the classifier. In this case, the generative process explores random and noisy configurations where the CNN has no training data, thus entering fake energy minima due to extrapolation.

\section{Discussion}


The Min-MaxEnt model is a novel generative AI algorithm that differs from most competitors by two significant features: (i) the approach stems from a solid theoretical apparatus rooted in fundamental physics and information theory, (ii) the entropic gradient update (Eq.~\ref{eq:minim:entropy}), that distills information into the observables~\cite{Lynn2025},  do not minimize  any metric distance between the generated samples and the training set.
As a consequence, the training process is never directly exposed to the training set, thus making it extremely hard for the model to store individual copies of the dataset and providing a better generalization in presence of rare data (see comparison with the VAE, \figurename~\ref{fig:4})). 
Indeed,  Min-MaxEnt models learn only generalized features/patterns across all the training sets. 
Moreover, the ability to control the output via discriminator networks trained \emph{a priori} promotes more effective interaction between the user and the generative process (see \figurename~\ref{fig:5} and \figurename~\ref{fig:6}), currently a significant limitation of most GenAI algorithms.
As shown in \figurename~\ref{fig:gan}, the Min-MaxEnt can bypass GenAI detection mechanisms without retraining the model by adding the detection function as a bias in the generative process. While this provides a systematic approach to enhancing the quality of synthetic data, in turn it may raise ethical concerns about the capability of  algorithmically distinguishing real from  generated content, which could have significant social implications. Therefore, it will be crucial for production applications built on this approach to address these concerns by incorporating watermarks or implementing mechanisms for identifying generated content.


In conclusion, Min-MaxEnt stands out as a first principles method to GenAI, offering a fundamentally different perspective to the field and paving the way to overcome current limitations of state-of-the-art approaches. 


\subsection*{Author contributions statement}
M.M and L.M. conceived the research, developed the model, wrote and revised the manuscript. 

\subsection*{Competing Interests statement}
The authors declare no competing financial or non-financial interests.


\bibliographystyle{unsrtnat}
\bibliography{mybib}

\begin{thebibliography}{35}
\providecommand{\natexlab}[1]{#1}
\providecommand{\url}[1]{\texttt{#1}}
\expandafter\ifx\csname urlstyle\endcsname\relax
  \providecommand{\doi}[1]{doi: #1}\else
  \providecommand{\doi}{doi: \begingroup \urlstyle{rm}\Url}\fi

\bibitem[Lv(2023)]{Lv2023}
Zhihan Lv.
\newblock Generative artificial intelligence in the metaverse era.
\newblock \emph{Cognitive Robotics}, 3:\penalty0 208–217, 2023.

\bibitem[Vaswani et~al.()Vaswani, Shazeer, Parmar, Uszkoreit, Jones, Gomez,
  Kaiser, and Polosukhin]{vaswani_attention_2023}
Ashish Vaswani, Noam Shazeer, Niki Parmar, Jakob Uszkoreit, Llion Jones,
  Aidan~N. Gomez, Lukasz Kaiser, and Illia Polosukhin.
\newblock Attention is all you need.
\newblock URL \url{http://arxiv.org/abs/1706.03762}.

\bibitem[Goodfellow et~al.(2020)Goodfellow, Pouget-Abadie, Mirza, Xu,
  Warde-Farley, Ozair, Courville, and Bengio]{Goodfellow2020}
Ian Goodfellow, Jean Pouget-Abadie, Mehdi Mirza, Bing Xu, David Warde-Farley,
  Sherjil Ozair, Aaron Courville, and Yoshua Bengio.
\newblock Generative adversarial networks.
\newblock \emph{Communications of the ACM}, 63\penalty0 (11):\penalty0
  139–144, October 2020.
\newblock ISSN 1557-7317.
\newblock \doi{10.1145/3422622}.
\newblock URL \url{http://dx.doi.org/10.1145/3422622}.

\bibitem[Kingma and Welling()]{kingma_auto-encoding_2022}
Diederik Kingma and Max Welling.
\newblock Auto-encoding variational bayes.

\bibitem[Song et~al.(2020)Song, Sohl-Dickstein, Kingma, Kumar, Ermon, and
  Poole]{diffmodel}
Yang Song, Jascha Sohl-Dickstein, Diederik~P. Kingma, Abhishek Kumar, Stefano
  Ermon, and Ben Poole.
\newblock Score-based generative modeling through stochastic differential
  equations.
\newblock 2020.

\bibitem[Gupta et~al.(2024)Gupta, Ding, Guan, and Ding]{Gupta2024}
Priyanka Gupta, Bosheng Ding, Chong Guan, and Ding Ding.
\newblock Generative ai: A systematic review using topic modelling techniques.
\newblock \emph{Data and Information Management}, 8\penalty0 (2):\penalty0
  100066, June 2024.

\bibitem[Wu et~al.(2024)Wu, Xia, Deng, Liu, Zhang, Guo, Cui, Pei, Wu, Xie,
  Chen, Lu, Hu, Wu, Chan, Chen, Zhou, Yu, Chen, Liu, Guo, Qin, and Liu]{Wu2024}
Kehan Wu, Yingce Xia, Pan Deng, Renhe Liu, Yuan Zhang, Han Guo, Yumeng Cui,
  Qizhi Pei, Lijun Wu, Shufang Xie, Si~Chen, Xi~Lu, Song Hu, Jinzhi Wu, Chi-Kin
  Chan, Shawn Chen, Liangliang Zhou, Nenghai Yu, Enhong Chen, Haiguang Liu,
  Jinjiang Guo, Tao Qin, and Tie-Yan Liu.
\newblock Tamgen: drug design with target-aware molecule generation through a
  chemical language model.
\newblock \emph{Nature Communications}, 15\penalty0 (1), October 2024.

\bibitem[Gangwal et~al.(2024)Gangwal, Ansari, Ahmad, Azad, Kumarasamy,
  Subramaniyan, and Wong]{Gangwal2024}
Amit Gangwal, Azim Ansari, Iqrar Ahmad, Abul~Kalam Azad, Vinoth Kumarasamy,
  Vetriselvan Subramaniyan, and Ling~Shing Wong.
\newblock Generative artificial intelligence in drug discovery: basic
  framework, recent advances, challenges, and opportunities.
\newblock \emph{Frontiers in Pharmacology}, 15, February 2024.

\bibitem[Trinquier et~al.(2021)Trinquier, Uguzzoni, Pagnani, Zamponi, and
  Weigt]{Trinquier2021}
Jeanne Trinquier, Guido Uguzzoni, Andrea Pagnani, Francesco Zamponi, and Martin
  Weigt.
\newblock Efficient generative modeling of protein sequences using simple
  autoregressive models.
\newblock \emph{Nature Communications}, 12\penalty0 (1), October 2021.

\bibitem[Zeni et~al.(2025)Zeni, Pinsler, Z\"{u}gner, Fowler, Horton, Fu, Wang,
  Shysheya, Crabbé, Ueda, Sordillo, Sun, Smith, Nguyen, Schulz, Lewis, Huang,
  Lu, Zhou, Yang, Hao, Li, Yang, Li, Tomioka, and Xie]{Zeni2025}
Claudio Zeni, Robert Pinsler, Daniel Z\"{u}gner, Andrew Fowler, Matthew Horton,
  Xiang Fu, Zilong Wang, Aliaksandra Shysheya, Jonathan Crabbé, Shoko Ueda,
  Roberto Sordillo, Lixin Sun, Jake Smith, Bichlien Nguyen, Hannes Schulz,
  Sarah Lewis, Chin-Wei Huang, Ziheng Lu, Yichi Zhou, Han Yang, Hongxia Hao,
  Jielan Li, Chunlei Yang, Wenjie Li, Ryota Tomioka, and Tian Xie.
\newblock A generative model for inorganic materials design.
\newblock \emph{Nature}, January 2025.

\bibitem[Totlani(2023)]{Totlani2023}
Ketan Totlani.
\newblock The evolution of generative ai: Implications for the media and film
  industry.
\newblock \emph{International Journal for Research in Applied Science and
  Engineering Technology}, 11\penalty0 (10):\penalty0 973–980, October 2023.

\bibitem[Law(2024)]{Law2024}
Locky Law.
\newblock Application of generative artificial intelligence (genai) in language
  teaching and learning: A scoping literature review.
\newblock \emph{Computers and Education Open}, 6:\penalty0 100174, June 2024.

\bibitem[Roumeliotis and Tselikas(2023)]{Roumeliotis2023}
Konstantinos~I. Roumeliotis and Nikolaos~D. Tselikas.
\newblock Chatgpt and open-ai models: A preliminary review.
\newblock \emph{Future Internet}, 15\penalty0 (6):\penalty0 192, May 2023.

\bibitem[Carlini et~al.(2023)Carlini, Hayes, Nasr, Jagielski, Sehwag,
  Tram{\`e}r, Balle, Ippolito, and Wallace]{carlini2023}
Nicolas Carlini, Jamie Hayes, Milad Nasr, Matthew Jagielski, Vikash Sehwag,
  Florian Tram{\`e}r, Borja Balle, Daphne Ippolito, and Eric Wallace.
\newblock Extracting training data from diffusion models.
\newblock In \emph{32nd USENIX Security Symposium (USENIX Security 23)}, pages
  5253--5270, Anaheim, CA, August 2023. USENIX Association.
\newblock ISBN 978-1-939133-37-3.
\newblock URL
  \url{https://www.usenix.org/conference/usenixsecurity23/presentation/carlini}.

\bibitem[Jones(2024)]{Jones2024}
Nicola Jones.
\newblock The ai revolution is running out of data. what can researchers do?
\newblock \emph{Nature}, 636\penalty0 (8042):\penalty0 290–292, December
  2024.
\newblock \doi{10.1038/d41586-024-03990-2}.

\bibitem[Shumailov et~al.(2024)Shumailov, Shumaylov, Zhao, Papernot, Anderson,
  and Gal]{Shumailov2024}
Ilia Shumailov, Zakhar Shumaylov, Yiren Zhao, Nicolas Papernot, Ross Anderson,
  and Yarin Gal.
\newblock Ai models collapse when trained on recursively generated data.
\newblock \emph{Nature}, 631\penalty0 (8022):\penalty0 755–759, July 2024.

\bibitem[Alemohammad et~al.(2023)Alemohammad, Casco-Rodriguez, Luzi, Humayun,
  Babaei, LeJeune, Siahkoohi, and Baraniuk]{poison_data}
Sina Alemohammad, Josue Casco-Rodriguez, Lorenzo Luzi, Ahmed~Imtiaz Humayun,
  Hossein Babaei, Daniel LeJeune, Ali Siahkoohi, and Richard~G. Baraniuk.
\newblock Self-consuming generative models go mad.
\newblock 2023.

\bibitem[Li et~al.()Li, Yang, Kuang, Wu, Wang, Xiao, and
  Chen]{li_controlnet_2024}
Ming Li, Taojiannan Yang, Huafeng Kuang, Jie Wu, Zhaoning Wang, Xuefeng Xiao,
  and Chen Chen.
\newblock {ControlNet}++: Improving conditional controls with efficient
  consistency feedback.
\newblock URL \url{http://arxiv.org/abs/2404.07987}.

\bibitem[Huang et~al.(2023)Huang, Chen, Liu, Shen, Zhao, and
  Zhou]{huang2023composer}
Lianghua Huang, Di~Chen, Yu~Liu, Yujun Shen, Deli Zhao, and Jingren Zhou.
\newblock Composer: Creative and controllable image synthesis with composable
  conditions.
\newblock \emph{arXiv preprint arXiv:2302.09778}, 2023.

\bibitem[Zhang et~al.()Zhang, Rao, and Agrawala]{zhang_adding_2023}
Lvmin Zhang, Anyi Rao, and Maneesh Agrawala.
\newblock Adding conditional control to text-to-image diffusion models.
\newblock URL \url{http://arxiv.org/abs/2302.05543}.

\bibitem[Hu et~al.()Hu, Shen, Wallis, Allen-Zhu, Li, Wang, Wang, and
  Chen]{hu_lora_2021}
Edward~J. Hu, Yelong Shen, Phillip Wallis, Zeyuan Allen-Zhu, Yuanzhi Li, Shean
  Wang, Lu~Wang, and Weizhu Chen.
\newblock {LoRA}: Low-rank adaptation of large language models.
\newblock URL \url{http://arxiv.org/abs/2106.09685}.

\bibitem[Xie et~al.(2023)Xie, Zhang, Lin, Hinz, and Zhang]{xie2023smartbrush}
Shaoan Xie, Zhifei Zhang, Zhe Lin, Tobias Hinz, and Kun Zhang.
\newblock Smartbrush: Text and shape guided object inpainting with diffusion
  model.
\newblock In \emph{Proceedings of the IEEE/CVF Conference on Computer Vision
  and Pattern Recognition}, pages 22428--22437, 2023.

\bibitem[Rombach et~al.(2022)Rombach, Blattmann, Lorenz, Esser, and
  Ommer]{rombach2022high}
Robin Rombach, Andreas Blattmann, Dominik Lorenz, Patrick Esser, and Bj{\"o}rn
  Ommer.
\newblock High-resolution image synthesis with latent diffusion models.
\newblock In \emph{Proceedings of the IEEE/CVF conference on computer vision
  and pattern recognition}, pages 10684--10695, 2022.

\bibitem[Bialek(2012)]{Bialek2012-lp}
William Bialek.
\newblock \emph{Biophysics}.
\newblock Princeton University Press, Princeton, NJ, October 2012.

\bibitem[Miotto and Monacelli(2018)]{ecolat}
Mattia Miotto and Lorenzo Monacelli.
\newblock Entropy evaluation sheds light on ecosystem complexity.
\newblock \emph{Phys. Rev. E}, 98:\penalty0 042402, Oct 2018.

\bibitem[Zhu et~al.(1997)Zhu, Wu, and Mumford]{6796444}
Song~Chun Zhu, Ying~Nian Wu, and David Mumford.
\newblock Minimax entropy principle and its application to texture modeling.
\newblock \emph{Neural Computation}, 9\penalty0 (8):\penalty0 1627--1660, 1997.

\bibitem[Lynn et~al.(2025)Lynn, Yu, Pang, Palmer, and Bialek]{Lynn2025}
Christopher~W. Lynn, Qiwei Yu, Rich Pang, Stephanie~E. Palmer, and William
  Bialek.
\newblock Exact minimax entropy models of large-scale neuronal activity.
\newblock \emph{Physical Review E}, 111\penalty0 (5), May 2025.

\bibitem[Miotto and Monacelli(2021)]{miotto_tolomeo_2021}
Mattia Miotto and Lorenzo Monacelli.
\newblock {TOLOMEO}, a {Novel} {Machine} {Learning} {Algorithm} to {Measure}
  {Information} and {Order} in {Correlated} {Networks} and {Predict} {Their}
  {State}.
\newblock \emph{Entropy}, 23\penalty0 (9):\penalty0 1138, September 2021.
\newblock Number: 9 Publisher: Multidisciplinary Digital Publishing Institute.

\bibitem[Kingma and Ba(2014)]{ADAM}
Diederik~P. Kingma and Jimmy Ba.
\newblock Adam: A method for stochastic optimization.
\newblock \emph{CoRR}, abs/1412.6980, 2014.
\newblock URL \url{https://api.semanticscholar.org/CorpusID:6628106}.

\bibitem[Ma and Dai(2011)]{Ma2011}
S.~Ma and Y.~Dai.
\newblock Principal component analysis based methods in bioinformatics studies.
\newblock \emph{Briefings in Bioinformatics}, 12\penalty0 (6):\penalty0
  714–722, January 2011.

\bibitem[Altsybeev(2020)]{Altsybeev2020}
I.~Altsybeev.
\newblock Application of principal component analysis to establish a proper
  basis for flow studies in heavy-ion collisions.
\newblock \emph{Physics of Particles and Nuclei}, 51\penalty0 (3):\penalty0
  314–318, May 2020.

\bibitem[Greenacre et~al.(2022)Greenacre, Groenen, Hastie, D’Enza, Markos,
  and Tuzhilina]{Greenacre2022}
Michael Greenacre, Patrick J.~F. Groenen, Trevor Hastie, Alfonso~Iodice
  D’Enza, Angelos Markos, and Elena Tuzhilina.
\newblock Principal component analysis.
\newblock \emph{Nature Reviews Methods Primers}, 2\penalty0 (1), December 2022.

\bibitem[Bredell et~al.(2023)Bredell, Flouris, Chaitanya, Erdil, and
  Konukoglu]{Bredell2023}
Gustav Bredell, Kyriakos Flouris, Krishna Chaitanya, Ertunc Erdil, and Ender
  Konukoglu.
\newblock Explicitly minimizing the blur error of variational autoencoders.
\newblock 2023.

\bibitem[E.~Alpaydin(1998)]{mnist}
C.~Kaynak E.~Alpaydin.
\newblock Optical recognition of handwritten digits.
\newblock 1998.
\newblock \doi{10.24432/C50P49}.
\newblock URL \url{https://archive.ics.uci.edu/dataset/80}.

\bibitem[Hastings(1970)]{Hastings1970}
W.~K. Hastings.
\newblock Monte carlo sampling methods using markov chains and their
  applications.
\newblock \emph{Biometrika}, 57\penalty0 (1):\penalty0 97–109, April 1970.

\end{thebibliography}

\end{document}